\pdfoutput=1
\documentclass[letterpaper]{article} 
\usepackage[preprint]{aaai2027}  
\usepackage[hyphens]{url}  
\usepackage{graphicx} 
\usepackage{natbib}  
\usepackage{caption} 
\usepackage{booktabs}
\usepackage{array}
\usepackage{multirow}
\usepackage{subcaption}
\usepackage{xspace}
\usepackage{amsmath}
\usepackage{amssymb}
\usepackage{algorithm}
\usepackage{algorithmic}
\usepackage{placeins}
\usepackage{xcolor}

\usepackage[most]{tcolorbox}
\usepackage{listings}
\usepackage{tabularx}

\lstdefinestyle{promptstyle}{
    basicstyle=\ttfamily\small,
    breaklines=true,
    breakatwhitespace=false,
    breakautoindent=false,
    breakindent=0pt,
    columns=fullflexible,
    keepspaces=true,
    showstringspaces=false,
    literate={–}{{\textrm{\textendash}}}1 {—}{{\textrm{\textemdash}}}1,
    frame=none
}

\newtcolorbox{promptbox}[1]{
    colback=gray!5,
    colframe=gray!50,
    title=\textbf{#1},
    fonttitle=\small,
    boxrule=0.4pt,
    arc=2pt,
    left=4pt,
    right=4pt,
    top=4pt,
    bottom=4pt,
    breakable
}

\newcommand{\OURS}{MT-SDPO\xspace}

\title{Learn from Whoever Is Right: Answer-Verified Multi-Teacher Distillation for Multi-Domain LLMs}

\author{
    Xixiang He\textsuperscript{\rm 1},
    Xingming Li\textsuperscript{\rm 1},
    Baiqi Wu\textsuperscript{\rm 2},
    Qiyao Sun\textsuperscript{\rm 1},
    Xuanyu Ji\textsuperscript{\rm 1},
    Ao Cheng\textsuperscript{\rm 1},
    Qingyong Hu\textsuperscript{\rm 3}\corresponding
}
\affiliations{
    \textsuperscript{\rm 1}National University of Defense Technology\\
    \textsuperscript{\rm 2}Zhejiang University\\
    \textsuperscript{\rm 3}Intelligent Game and Decision Lab\\
    \{hexixiang, lixingming, sunqiyao18, jixuanyu18, chengao18\}@nudt.edu.cn,\\
    wubaiqi@zju.edu.cn, huqingyong15@outlook.com
}

\begin{document}

\maketitle

\begin{abstract}
Modern large language models (LLMs) rely on reinforcement learning to build strong capabilities in individual domains, but integrating those capabilities into a single deployable model remains challenging. By routing each sample to the teacher whose domain matches it, existing approaches let a domain label decide which teacher provides supervision. However, domain expertise holds only on average: the matched teacher is not always correct on a given sample, while a teacher from another domain sometimes is. The reliable teacher therefore has to be identified per sample, not per domain. In this paper, we introduce \textbf{M}ulti-\textbf{T}eacher \textbf{S}elf-\textbf{D}istillation \textbf{P}olicy \textbf{O}ptimization (\OURS), an on-policy distillation method that unifies several frozen teachers into one student model. \OURS consists of three components: (1) \emph{self-anchors}, where a rollout is supervised by a correct rollout from its own group; (2) \emph{answer-verified eligibility}, where a teacher may supervise a sample only if its own answer passes a verifier; and (3) \emph{privileged distillation}, which merges the anchor and all verified feedback into one context that an exponential moving average self-teacher reads and the student does not, thereby keeping one policy at deployment. Across five students from three model families, \OURS lifts the weakest domain of Qwen3-8B by 14.79 points and narrows its domain gap by 74.7\%, a better balance than serving one matched teacher per domain. Verified reliability, not domain membership, should decide who teaches. Code is available at \url{https://github.com/hexixiang/MT-SDPO}.
\end{abstract}

\section{Introduction}
\label{sec:intro}

\begin{quote}
    \itshape ``When three walk together, there is always a teacher among them: I choose what is good and follow it; what is not good, I correct.''

    \hfill --- Confucius, \emph{The Analects}
\end{quote}

Post-training a modern large language model (LLM) is organized by domain: verifiable-answer reinforcement learning (RL) for mathematical reasoning~\citep{shao2024deepseekmath,guo2025deepseekr1}, software-evolution rewards~\citep{wei2025swerl} and executable SWE environments~\citep{jain2025r2e}, human-preference rewards for instruction following~\citep{ouyang2022instructgpt}, and interactive environments for search agents~\citep{nakano2021webgpt,jin2025searchr1}. Each recipe needs its own data and rollout protocol, so a separate domain teacher is trained for each domain. A deployed system, however, answers every request with one policy: a teacher pool adds overhead and requests cross domain boundaries. Multi-domain post-training therefore ends by asking how to fold the teachers back into one deployable model.

Among the routes to this problem, we study on-policy integration, which keeps the teachers frozen and lets one student learn on its own rollouts, so what remains open is which teacher supervises which rollout. Multi-Teacher On-Policy Distillation (MOPD)~\citep{mopd2026} develops domain teachers independently, then routes each rollout to the teacher whose domain matches the prompt, assuming that teacher is right for every sample in its domain.

\begin{figure}[t]
    \centering
    \includegraphics[width=\columnwidth]{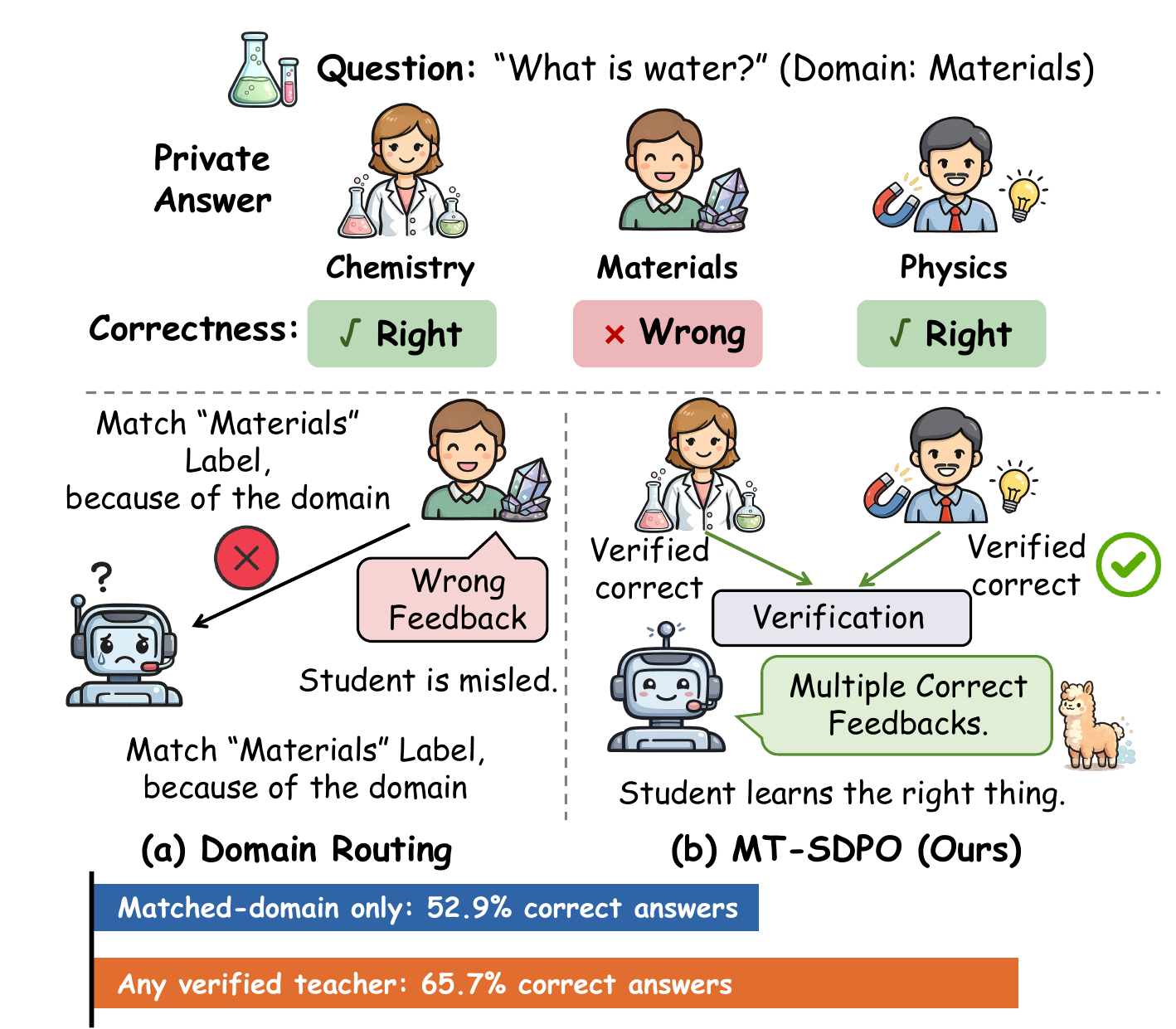}
    \caption{Sample-Level Teacher Reliability Gap. A domain label can select a teacher wrong on the sample; \OURS verifies private answers and uses all eligible feedback.}
    \label{fig:teaser}
\end{figure}

Domain expertise, however, is an average-case property: domain boundaries overlap, teachers trained from a shared source repeat the same errors, and a teacher from another domain sometimes succeeds anyway. Figure~\ref{fig:teaser} quantifies this gap on our training set: the matched teacher is correct on only 52.9\% of samples, while per-sample selection raises the share with at least one correct teacher to 65.7\%. The reliable teacher therefore varies~per~sample, not~per~domain: \textbf{whom should the student trust?}

We answer with an allocation rule that drops the domain label: on each sample, any source whose answer is verified correct is a legitimate teacher, and no unverified source is. The student counts as one such source: its own correct rollout is the cheapest verified teacher, so frozen teachers are consulted only where it fails. Supervision then flows exactly where someone is known to be right.

\begin{figure*}[t]
    \centering
    \includegraphics[width=0.97\textwidth]{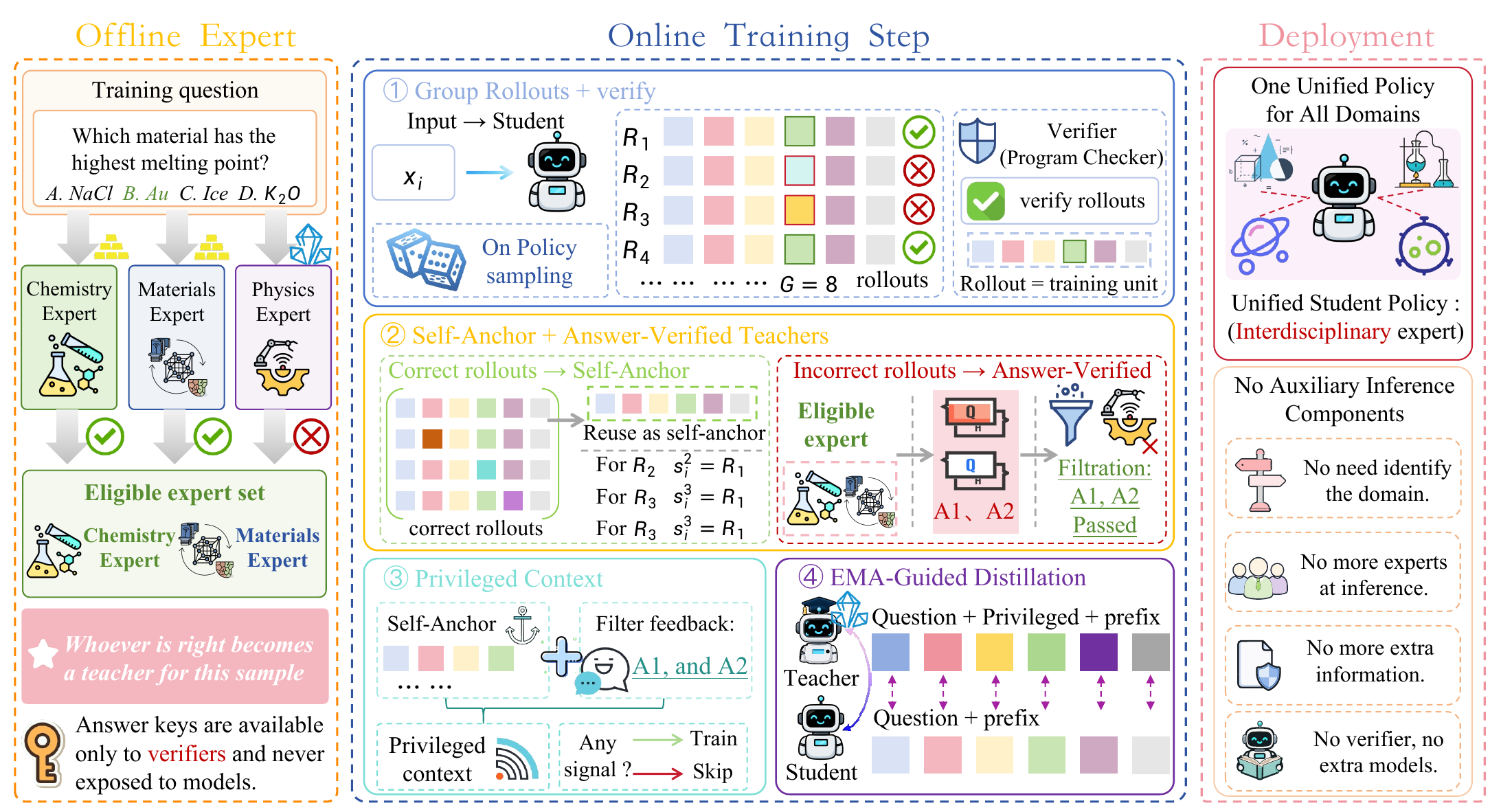}
    \caption{Overview of \OURS. Offline answer verification builds a sample-level eligible teacher set. Online, self-anchors and sanitized teacher feedback condition an EMA self-teacher that distills into the student model, all that deployment retains.}
    \label{fig:method-overview}
\end{figure*}

We instantiate the rule as \textbf{Multi-Teacher Self-Distillation Policy Optimization (\OURS)} on top of Self-Distillation Policy Optimization (SDPO)~\citep{sdpo}, which turns textual feedback into a dense token-level signal through a privileged self-teacher. Three technical questions remain, and one component answers each: anchoring where the student is already right, coverage where it is not and the anchor is missing, and merging where anchor and feedback must become one signal. Figure~\ref{fig:method-overview} traces one training step through the three components. Self-anchors let a correct rollout supervise the incorrect peers of its own group. Answer-verified eligibility admits a teacher's feedback only when its private answer passes a verifier that no model sees, and takes every teacher that qualifies. Privileged distillation merges the anchor and all verified feedback into one context that only an exponential moving average (EMA) self-teacher reads, and distills that self-teacher into the student token by token. Training updates only the student model, which is also all that deployment keeps.

Built on per-sample answer verification and privileged self-distillation, \OURS not only redistributes capability toward the weakest domain, but can also outperform a three-model reference that serves one matched teacher per domain. We evaluate on the chemistry, materials science, and physics domains of the Science Q\&A subset in SciKnowEval L3~\citep{sciknoweval} with five student models from three families, each under its own multi-domain initialization. On Qwen3-8B~\citep{qwen2025qwen3}, \OURS gains 4.64 points in Macro accuracy and 14.79 in worst-domain accuracy, and cuts the domain gap from 20.96 to 5.30. The gains are not tied to one model: every Qwen3 scale becomes more balanced, though the aggregate improves only at 8B, and all three metrics improve together on OLMo-3-7B-Instruct~\citep{olmo3}, a second family with a different teacher generator. On Llama-3.1-8B-Instruct~\citep{grattafiori2024llama3}, whose initialization is already balanced and leaves little headroom, no online method pays off, a negative result. Our contributions are as follows:

\begin{itemize}
    \item \textbf{Sample-level diagnosis.} We quantify the reliability gap between domain-matched and per-sample teacher selection, separating matched coverage, cross-domain rescue, and failure across all teachers.
    \item \textbf{\OURS.} Answer-verified multi-teacher self-distillation that grants eligibility per sample and consolidates all eligible teachers with the student's self-anchor into one privileged self-teacher.
    \item \textbf{Controlled evidence.} Same-initialization ablations isolating answer verification, per-sample selection, and multi-teacher aggregation.
\end{itemize}

\section{Related Work}
\label{sec:related}

Turning separately trained domain teachers into one deployable policy raises three questions that organize prior work: how the teachers are consolidated into one model, how a student learns on-policy from a teacher, and how supervision is allocated across teachers on a shared sample.

\textbf{(1) Capability Consolidation for Multi-Domain LLMs.}
Prior routes build or consolidate multi-domain capabilities by 1) joint or cascaded multi-domain RL~\citep{qwen2025qwen3, wang2026nemotroncascadescalingcascadedreinforcement}, 2) off-policy finetuning on teacher completions~\citep{deepseek2025v32}, 3) weight-space merging~\citep{wortsman2022soups, ilharco2023arithmetic}, or 4) mixed-domain training with negative-transfer control~\citep{yu2020gradient, cai2026advancing, yang2026disentangling, ye2026synergy}. Closest to us is MOPD~\citep{mopd2026}, which routes the student's own rollouts to the domain-matched frozen teacher; we implement this rule as our \emph{Domain Routing} baseline. It assumes the matched teacher is the right supervisor for every sample from its domain, whereas \OURS keeps MOPD's two-stage split but selects the supervisor per sample by verification.

\textbf{(2) On-Policy and Self-Distillation.}
Classical distillation matches a teacher on a fixed corpus and is off-policy, so the student can drift at inference~\citep{hinton2015distilling, kim2016seqkd}. Adaptive off-policy distillation instead reuses student-generated outputs~\citep{ko2024distillm}, while on-policy distillation scores the student's current rollouts with a separate teacher~\citep{gu2024minillm, agarwal2023onpolicy}. Self- and context-distillation let a model learn from signals produced by itself or by a context-privileged copy~\citep{sdpo, opsd, kim2026rebellious, song2026expanding, ye2026policy}. Our baseline SDPO~\citep{sdpo} anchors this self-teacher on the student's own correct rollout, and gives no signal where that rollout is missing. \OURS keeps the self-anchor but adds verified external teachers, so those unanchored samples can still be supervised by whichever teacher is right.

\textbf{(3) Multi-Teacher Distillation and Supervision Reliability.}
Distilling several teachers into one student~\citep{you2017learning} appears in recent LLMs as output mixing or per-sample routing~\citep{xiaomi2026mimov2flash, yang2026nemotron}, sometimes with adaptive per-instance teacher weights~\citep{liu2020adaptive, yuan2021reinforced}. Separately, verifiable-reward RL and generative verifiers turn correctness into a training signal that scores the student's own output~\citep{lambert2024tulu, guo2025deepseekr1, zhang2024generative}; \OURS draws on the same signal but spends it on deciding whom the student should trust. However, the multi-teacher aggregation and routing methods cited above do not use the correctness of each teacher's private answer to gate supervision; an incorrect teacher can therefore be weighted or routed like a correct one. \OURS closes this gap and differs in three ways: 1) a teacher may supervise a sample only if its own private answer passes a verifier, so eligibility reflects correctness rather than a domain label or learned weight; 2) eligibility is decided per sample, so a cross-domain teacher helps whenever it is right; 3) all verified teachers combine with the student's self-anchor into a single self-teacher.

\section{Preliminaries}
\label{sec:prelim}

\paragraph{Self-Distillation Policy Optimization.}
Given a prompt $x\sim\mathcal D$, SDPO~\citep{sdpo} samples an on-policy rollout $y=(y_1,\ldots,y_T)\sim\pi_\theta(\cdot\mid x)$ and re-evaluates it with a self-teacher that additionally observes privileged information $c$. Write $p_{\theta,t}=\pi_\theta(\cdot\mid x,y_{<t})$ for the student distribution and $q_{\bar\theta,t}=\operatorname{sg}[\pi_{\bar\theta}(\cdot\mid x,c,y_{<t})]$ for the self-teacher distribution, where $\bar\theta$ is a slowly updated copy of $\theta$. The rollout-level objective is
\begin{equation}
\mathcal L_{\mathrm{SDPO}}(x,y,c)
=\frac{1}{T}\sum_{t=1}^{T}D\!\left(p_{\theta,t}\middle\|q_{\bar\theta,t}\right),
\label{eq:sdpo-objective}
\end{equation}
where $D$ is a token-level divergence and $\operatorname{sg}$ blocks self-teacher gradients. The context $c$ can contain environment feedback or another verified successful rollout; it changes the self-teacher distribution without changing the rollout on which the student is trained.

\paragraph{Multi-Teacher Setting.}
Each training instance consists of a prompt $x_i$, a verifiable reference answer $a_i^*$, and a domain label $d_i$. We train one student model $\pi_\theta$ with feedback from $K$ frozen domain teachers $\{E_k\}_{k=1}^{K}$, each queried either for a private answer, written $E_k^{\mathrm{ans}}$, or for a critique of a given rollout, written $E_k^{\mathrm{fb}}$. A programmatic verifier $\textsc{Verify}(\cdot,a_i^*)\in\{0,1\}$ extracts the answer from a completion and compares it with $a_i^*$; the reference answer reaches this verifier and the sanitizer only, and never enters a student or teacher prompt at any stage of \OURS. Our problem is to decide, for each student rollout, which teachers may contribute to $c$ and then consolidate their supervision into the single policy $\pi_\theta$ that training retains.

\section{Method}
\label{sec:method}

We propose \OURS, an answer-verified multi-teacher distillation framework that consolidates $K$ frozen domain teachers into one deployable policy. A single allocation rule governs it: a teacher may supervise a sample only if its own private answer is verified correct. \OURS therefore pairs a supervision source the student produces itself with a \emph{sample-level eligible teacher set} that replaces domain routing.

\subsection{Framework Overview}
\label{sec:framework-overview}

\paragraph{Verification runs once, before training.} As illustrated in Figure~\ref{fig:method-overview}, \OURS trains a single student model $\pi_\theta$ with three auxiliary parts that all disappear at inference: $K$ frozen domain teachers, a programmatic verifier, and an EMA copy $\pi_{\bar\theta}$ of the student that serves as the self-teacher. Every teacher first answers each question privately, and the verifier caches the correct ones as that sample's eligible set, so training pays no recurring answering cost.

\paragraph{Each online step aligns supervision to individual rollouts.} The student samples a group of rollouts per prompt and the verifier labels them. A correct rollout supervises the remaining rollouts of its group as their self-anchor, whereas an incorrect one is sent to the eligible teachers, whose sanitized feedback joins the anchor in a privileged context that only the self-teacher reads. Distilling the self-teacher into the student is the only parameter update.

\subsection{Rollout-Specific Self-Anchors}
\label{sec:self-anchor}

The first source of privileged information is the student itself~\citep{sdpo}: a rollout group usually mixes correct and incorrect rollouts, and a correct one is a solution the current policy has already shown it can produce.

\paragraph{A programmatic check splits each rollout group.} For prompt $x_i$, the student samples $G$ rollouts $\{y_i^1,\ldots,y_i^G\}\sim\pi_\theta(\cdot\mid x_i)$, and
\begin{equation}
\mathcal S_i=\left\{j:\textsc{Verify}(y_i^j,a_i^*)=1\right\}
\label{eq:correct-set}
\end{equation}
indexes the correct ones.

\paragraph{Each rollout takes its self-anchor from a correct peer.} The self-anchor of rollout $j$ is
\begin{equation}
s_i^j=
\begin{cases}
\operatorname{Trunc}\!\left(y_i^{j^\star}\right), & \mathcal S_i\setminus\{j\}\neq\varnothing,\\
\varnothing, & \text{otherwise},
\end{cases}
\label{eq:self-anchor}
\end{equation}
where $j^\star=\min(\mathcal S_i\setminus\{j\})$ is the earliest correct rollout of the same group other than $j$, and $\operatorname{Trunc}$ keeps that rollout's solution segment within the anchor token budget.

\paragraph{Excluding the rollout itself prevents sequence leakage.} If $j$ were its own anchor, the self-teacher would observe a complete copy of the sequence whose prefixes it scores, and the supervision would degenerate into copying. An incorrect rollout can still reuse a correct peer, whereas a prompt that the student never solves has no self-anchor at all.

\subsection{Answer-Verified Teacher Eligibility}
\label{sec:teacher-eligibility}

Domain match indicates which teacher is stronger on average, not which is correct on a given question~\citep{mopd2026}. \OURS replaces this proxy with a per-sample check, and spends the resulting teacher queries only on rollouts the verifier marked incorrect, which is also where the self-anchor may be missing.

\paragraph{A private answer decides eligibility before training starts.} Teacher $k$ independently produces a private completion $\hat y_{i,k}=E_k^{\mathrm{ans}}(x_i)$ from the question and its choice list, and the verifier keeps the teachers that are correct:
\begin{equation}
\mathcal V_i=\left\{k:\textsc{Verify}(\hat y_{i,k},a_i^*)=1\right\}.
\label{eq:eligible-set}
\end{equation}
The eligible set is sample-specific and ignores the domain label $d_i$: a teacher from another domain enters $\mathcal V_i$ whenever it answers correctly, and the nominal-domain teacher drops out whenever it does not.

\paragraph{Every eligible teacher is queried independently.} Each $k\in\mathcal V_i$ receives the question, the full choice list, and the incorrect rollout, but neither $a_i^*$ nor $\hat y_{i,k}$. Only the messages that survive sanitization reach the context, concatenated in a fixed teacher order:
\begin{equation}
F_i^j=
\begin{cases}
\operatorname{Concat}_{k\in\mathcal V_i}\left[\Gamma\!\left(E_k^{\mathrm{fb}}(x_i,y_i^j)\right)\right], & j\notin\mathcal S_i,\\
\varnothing, & \text{otherwise},
\end{cases}
\label{eq:feedback-set}
\end{equation}
where the sanitizer $\Gamma$ screens a message against $x_i$ and $a_i^*$ and returns either the accepted feedback or $\varnothing$, so that an empty $\mathcal V_i$ also leaves $F_i^j$ empty. Eligibility fixes who is asked rather than who contributes: $|\mathcal V_i|$ eligible teachers supply at most $|\mathcal V_i|$ messages. Unlike conventional multi-teacher aggregation~\citep{you2017learning}, \OURS applies neither a majority vote nor domain-based weighting over those messages.

\paragraph{The sanitizer bounds label leakage and topical drift.} $\Gamma$ rejects an entire message that reveals the correct label, whether as a structured answer pattern, a targeted label expression, or the full text of the correct option. It also requires a short concept anchor occurring verbatim in the question, so that the critique addresses the specific error. Both gates are lexical: they bound leakage and drift, but neither catch every paraphrase of the label nor certify that the retained feedback is semantically valid.

\paragraph{Verification is an allocation rule, not a correctness proof.} Equation~\eqref{eq:eligible-set} tests one cached completion under a fixed prompt and decoding configuration; it neither tests the critique that the teacher writes later nor establishes general competence on the question. We therefore justify it by its downstream effect, which Section~\ref{sec:ablations} measures against domain routing and against unverified aggregation.

\subsection{Privileged Distillation and Student Update}
\label{sec:multi-teacher-update}

\OURS merges the two sources into one training signal: it conditions the EMA self-teacher on both and transfers the resulting token distribution to the student, which observes neither.

\paragraph{The privileged context combines the two sources under one mask.} For each rollout, \OURS forms the context $c_i^j=[s_i^j;F_i^j]$, which the self-teacher reads after the original question under natural-language section headers, together with the eligibility mask
\begin{equation}
m_i^j=\mathbb 1\!\left[c_i^j\neq\varnothing\right]=\mathbb 1\!\left[s_i^j\neq\varnothing\ \lor\ F_i^j\neq\varnothing\right].
\label{eq:privileged-mask}
\end{equation}
The context has four cases: full conditioning when both sources exist, a fallback to self-anchor SDPO~\citep{sdpo} when only $s_i^j$ exists, and teacher-only SDPO when only $F_i^j$ exists. The mask collapses the fourth, in which both are empty because the group holds no other correct rollout and no message survived, so that such a rollout contributes no distillation loss instead of an uninformative one. An empty source deletes its own block rather than leaving a placeholder, so the context stays a single natural-language prompt throughout. Appendix~\ref{sec:appendix-prompts} reproduces the template.

\paragraph{The objective is a masked, importance-weighted token-level divergence.} At token $t$ of rollout $y_i^j$, the student scores the sampled prefix from the question alone, $p_{\theta,i,t}^{j}=\pi_\theta(\cdot\mid x_i,y_{i,<t}^j)$, while the self-teacher scores the same prefix with the context attached, $q_{\bar\theta,i,t}^{j}=\operatorname{sg}[\pi_{\bar\theta}(\cdot\mid x_i,c_i^j,y_{i,<t}^j)]$. The teacher branch carries the EMA parameters $\bar\theta$ rather than the current $\theta$, and $\operatorname{sg}$ only keeps the update out of that branch. Teacher forcing on a shared prefix keeps the two branches token-aligned. Extending Eq.~\eqref{eq:sdpo-objective} from one rollout to a masked, importance-weighted batch objective and instantiating $D$ as the symmetric Jensen--Shannon divergence gives
\begin{equation}
\mathcal L_{\mathrm{MT\text{-}SDPO}}=\frac{1}{Z}\sum_{i,j,t}
m_i^jM_{i,t}^{j}w_{i,t}^{j}\,
D_{\mathrm{JSD}}\!\left(p_{\theta,i,t}^{j}\,\middle\Vert\,q_{\bar\theta,i,t}^{j}\right),
\label{eq:mt-sdpo-loss}
\end{equation}
where $M_{i,t}^{j}$ masks response tokens, $w_{i,t}^{j}$ is a detached and clipped importance weight~\citep{schulman2017ppo} that corrects policy drift on the sampled token, and $Z$ counts the unmasked tokens. We adopt the symmetric divergence and the EMA self-teacher because both stabilize on-policy distillation~\citep{agarwal2023onpolicy,sdpo}. Since $\pi_{\bar\theta}$ is both an EMA copy and the only branch that reads $c_i^j$, the divergence transfers a context-conditioned and temporally smoothed shift rather than the isolated contribution of $c_i^j$. Appendix~\ref{sec:appendix-online-details} gives the truncated support of the divergence, the two ratios behind $w_{i,t}^{j}$, and the per-micro-batch normalization that replaces $Z$.

\paragraph{Only the student model is updated and deployed.} Gradients reach $\theta$ alone: every $E_k$ stays frozen, and the EMA copy runs without gradients and follows $\bar\theta\leftarrow(1-\tau)\bar\theta+\tau\theta$ with EMA rate $\tau=0.05$ once per collected rollout batch. A step samples rollouts from $\pi_\theta$, computes every target in Eq.~\eqref{eq:mt-sdpo-loss} from one fixed $\bar\theta$, takes the mini-batch updates on $\theta$, and refreshes $\bar\theta$ last. Although the trainer reuses Group Relative Policy Optimization (GRPO)~\citep{shao2024deepseekmath} infrastructure to collect rollouts, no scalar advantage enters Eq.~\eqref{eq:mt-sdpo-loss}: the binary check only marks success, builds self-anchors, and gates teacher queries. Algorithm~\ref{alg:mt-sdpo} in Appendix~\ref{sec:appendix-online-details} states one full training step. Once training finishes, \OURS discards the EMA state and the teacher pool.

\section{Experiments}
\label{sec:experiments}

We ask three questions. \textbf{Q1: Supervision allocation.} Does a prompt's domain identify a teacher correct on it? \textbf{Q2: Capability integration.} Can \OURS produce one policy strong in aggregate and its weakest domain? \textbf{Q3: Component effects.} How much do answer verification, cross-domain reassignment, and aggregation each contribute?

\begin{figure}[t]
    \centering
    \captionsetup[subfigure]{font=small,labelfont=normalfont,textfont=normalfont}

    \begin{subfigure}[t]{0.56\columnwidth}
        \centering
        \includegraphics[width=\textwidth]{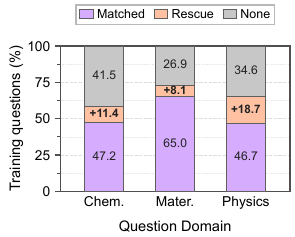}
        \caption{Teacher Coverage}
    \end{subfigure}
    \hfill
    \begin{subfigure}[t]{0.42\columnwidth}
        \centering
        \includegraphics[width=\textwidth]{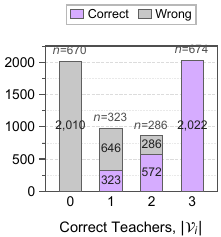}
        \caption{Answer Composition}
    \end{subfigure}

    \caption{Teacher-Pool Diagnosis on the Training Set. (a) Coverage from matched teachers, cross-domain rescue, and none correct. (b) Correct and wrong teacher answers per correct-teacher bin; labels give sample counts.}
    \label{fig:expert-pool-diagnostic}
\end{figure}

\begin{figure}[t]
    \centering
    \captionsetup[subfigure]{font=small,labelfont=normalfont,textfont=normalfont}

    \begin{subfigure}[t]{0.5125\columnwidth}
        \centering
        \includegraphics[width=\textwidth]{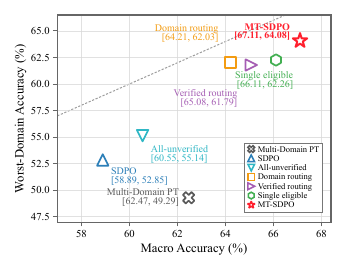}
        \caption{Absolute Accuracy}
        \label{fig:main-endpoint-absolute}
    \end{subfigure}
    \hfill
    \begin{subfigure}[t]{0.4775\columnwidth}
        \centering
        \includegraphics[width=\textwidth]{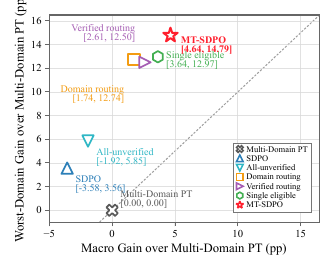}
        \caption{Accuracy Gain}
        \label{fig:main-endpoint-gain}
    \end{subfigure}

    \caption{Step-100 Capability Integration on Qwen3-8B. (a) Macro against Worst-domain accuracy. (b) Gains over Multi-Domain Post-Training. Dashed diagonals mark parity.}
    \label{fig:main-endpoint}
\end{figure}

\begin{table*}[t]
\centering
\small
\setlength{\tabcolsep}{3pt}
\renewcommand{\arraystretch}{1.0}
\newcommand{\tabledelta}[1]{\mbox{\,$(#1)$}}
\begin{tabular*}{\textwidth}{@{\extracolsep{\fill}}ll>{\raggedleft\arraybackslash}p{13.5mm}*{2}{>{\raggedleft\arraybackslash}p{12mm}}*{3}{>{\raggedleft\arraybackslash}p{21mm}}@{}}
\toprule
Model & Method & Chemistry & Materials & Physics & Macro $\uparrow$ ($\Delta$) & Worst $\uparrow$ ($\Delta$) & Gap $\downarrow$ ($\Delta$) \\
\midrule
\multirow{6}{*}{Qwen3-1.7B}
 & Base & 37.10 & 41.14 & 46.52 & 41.59 & 37.10 & 9.41 \\
 & Matched Teachers (3) & 45.33 & 65.59 & 50.24 & 53.72 & 45.33 & 20.25 \\
 & Multi-Domain PT & 42.33 & 63.05 & 51.66 & 52.35 & 42.33 & 20.73 \\
\cmidrule(lr){2-8}
 & SDPO & 40.74 & 55.93 & 50.40 & 49.02\tabledelta{-3.33} & 40.74\tabledelta{-1.59} & 15.19\tabledelta{-5.54} \\
 & Domain Routing & 45.33 & 55.46 & 44.78 & 48.52\tabledelta{-3.83} & 44.78\tabledelta{+2.45} & 10.68\tabledelta{-10.05} \\
 & \OURS & 46.52 & 56.65 & 46.20 & \textbf{49.79}\tabledelta{-2.56} & \textbf{46.20}\tabledelta{+3.87} & \textbf{10.44}\tabledelta{-10.29} \\
\midrule
\multirow{6}{*}{Qwen3-4B}
 & Base & 41.69 & 58.94 & 59.65 & 53.43 & 41.69 & 17.96 \\
 & Matched Teachers (3) & 51.27 & 69.38 & 66.06 & 62.24 & 51.27 & 18.12 \\
 & Multi-Domain PT & 48.66 & 71.36 & 68.67 & 62.90 & 48.66 & 22.71 \\
\cmidrule(lr){2-8}
 & SDPO & 51.98 & 70.17 & 64.40 & \textbf{62.18}\tabledelta{-0.72} & \textbf{51.98}\tabledelta{+3.32} & 18.20\tabledelta{-4.51} \\
 & Domain Routing & 47.78 & 63.77 & 65.27 & 58.94\tabledelta{-3.96} & 47.78\tabledelta{-0.88} & 17.48\tabledelta{-5.23} \\
 & \OURS & 50.63 & 62.26 & 63.92 & 58.94\tabledelta{-3.96} & 50.63\tabledelta{+1.97} & \textbf{13.29}\tabledelta{-9.42} \\
\midrule
\multirow{6}{*}{Qwen3-8B}
 & Base & 44.38 & 58.47 & 66.22 & 56.36 & 44.38 & 21.84 \\
 & Matched Teachers (3) & 52.93 & 68.04 & 68.99 & 63.32 & 52.93 & 16.06 \\
 & Multi-Domain PT & 49.29 & 70.25 & 67.88 & 62.47 & 49.29 & 20.96 \\
\cmidrule(lr){2-8}
 & SDPO & 54.11 & 69.70 & 52.85 & 58.89\tabledelta{-3.58} & 52.85\tabledelta{+3.56} & 16.85\tabledelta{-4.11} \\
 & Domain Routing & 62.03 & 67.33 & 63.29 & 64.21\tabledelta{+1.74} & 62.03\tabledelta{+12.74} & \textbf{5.30}\tabledelta{-15.66} \\
 & \OURS & 67.88 & 69.38 & 64.08 & \textbf{67.11}\tabledelta{+4.64} & \textbf{64.08}\tabledelta{+14.79} & \textbf{5.30}\tabledelta{-15.66} \\
\midrule
\multirow{6}{*}{\shortstack[l]{Llama-3.1\\8B-Instruct}}
 & Base & 30.38 & 43.04 & 41.53 & 38.32 & 30.38 & 12.66 \\
 & Matched Teachers (3) & 62.03 & 63.69 & 56.57 & 60.76 & 56.57 & 7.12 \\
 & Multi-Domain PT & 56.33 & 59.57 & 51.42 & 55.78 & 51.42 & 8.15 \\
\cmidrule(lr){2-8}
 & SDPO & 58.62 & 58.86 & 39.24 & 52.24\tabledelta{-3.54} & 39.24\tabledelta{-12.18} & 19.62\tabledelta{+11.47} \\
 & Domain Routing & 57.20 & 53.96 & 43.43 & 51.53\tabledelta{-4.25} & 43.43\tabledelta{-7.99} & \textbf{13.77}\tabledelta{+5.62} \\
 & \OURS & 59.57 & 53.48 & 44.62 & \textbf{52.56}\tabledelta{-3.22} & \textbf{44.62}\tabledelta{-6.80} & 14.95\tabledelta{+6.80} \\
\midrule
\multirow{6}{*}{\shortstack[l]{OLMo-3\\7B-Instruct}}
 & Base & 36.16 & 57.83 & 62.82 & 52.27 & 36.16 & 26.66 \\
 & Matched Teachers (3) & 48.02 & 61.55 & 53.24 & 54.27 & 48.02 & 13.53 \\
 & Multi-Domain PT & 48.42 & 63.21 & 57.67 & 56.43 & 48.42 & 14.79 \\
\cmidrule(lr){2-8}
 & SDPO & 45.41 & 62.66 & 54.91 & 54.33\tabledelta{-2.10} & 45.41\tabledelta{-3.01} & 17.25\tabledelta{+2.46} \\
 & Domain Routing & 51.90 & 64.56 & 63.29 & 59.92\tabledelta{+3.49} & 51.90\tabledelta{+3.48} & 12.66\tabledelta{-2.13} \\
 & \OURS & 55.69 & 65.82 & 64.55 & \textbf{62.02}\tabledelta{+5.59} & \textbf{55.69}\tabledelta{+7.27} & \textbf{10.13}\tabledelta{-4.66} \\
\bottomrule
\end{tabular*}
\caption{Capability Integration Across Five Student Models. Per-domain columns are avg@16 accuracy (\%) at step 100. Matched Teachers (3) is a per-domain frozen-teacher reference, not a deployable policy; Multi-Domain PT initializes the three online rows below it. $\Delta$ is each online row minus its same-model initialization, and bold marks the best aggregate per block.}
\label{tab:main-results}
\end{table*}

\subsection{Experimental Setup}

\paragraph{Datasets and Models.} We use the Science Q\&A subset of the L3 split of SciKnowEval~\citep{sciknoweval}, restricted to its chemistry, materials science, and physics domains, whose multiple-choice questions carry the single reference option that the verifier checks. We balance the three domains and construct an approximately 9:1 train--test split. The students are Qwen3-1.7B, Qwen3-4B, Qwen3-8B~\citep{qwen2025qwen3}, Llama-3.1-8B-Instruct~\citep{grattafiori2024llama3}, and OLMo-3-7B-Instruct~\citep{olmo3}, each supervised on the same prompts by a family-matched generator: Qwen3-32B, Llama-3.1-70B-Instruct, or OLMo-3.1-32B-Instruct. Domain-specific post-training gives every student three frozen teachers; multi-domain post-training on the union of the three training sets gives the shared initialization of all controlled online methods.

\paragraph{Protocol and Metrics.} Every controlled online method starts from that initialization and trains for 100 steps with 8 on-policy rollouts per prompt on 8 H100 80GB GPUs. Evaluation draws 16 responses for each of the 79 held-out samples per domain and reports avg@16 domain accuracy $A_d$, from which Macro, Worst, and Gap are the mean, minimum, and range over the three domains. We compare all methods at step 100 and measure every point difference against the same-student Multi-Domain Post-Training policy.

\paragraph{Main Baselines.} \textbf{SDPO}~\citep{sdpo} spends the same verifier entirely on the student's own rollouts, promoting successful ones to privileged self-supervision, so it is the teacher-free control for what external feedback adds. \textbf{Domain Routing} implements MOPD's domain-assignment rule~\citep{mopd2026}, sending every failed rollout to its nominal-domain teacher without checking whether it can solve the sample. It inherits the self-anchor, sanitizer, and privileged self-teacher, so only the routing rule differs.

\subsection{Domain Match Misses Correct Teachers}
\label{sec:expert-pool-diagnosis}

\paragraph{Domain match holds on average, not per sample.} Before training, we record the eligible set $\mathcal V_i$ of Eq.~\eqref{eq:eligible-set} on every training sample. Each teacher does improve its own held-out domain over the shared base, by 8.55, 9.57, and 2.77 points. Per sample they are far less separable: the matched teacher is correct on only 52.94\% of samples, and on physics samples the chemistry teacher is the more accurate of the two. Teachers sharing one base model and one data pipeline overlap in competence more than their labels suggest.

\paragraph{Per-sample selection turns that overlap into coverage.} Admitting any teacher whose private answer is verified correct raises coverage to 65.69\% and recovers 27.09\% of what the matched teacher misses. Figure~\ref{fig:expert-pool-diagnostic}(a) breaks this rescue down by domain: largest in physics, at 18.74 points, where domain match is weakest. Figure~\ref{fig:expert-pool-diagnostic}(b) regroups the samples by number of correct teachers: roughly half admit two or three, making aggregation possible, while 34.31\% admit none and bound any method built on this pool. That group is why \OURS pairs verified feedback with self-anchors.

\subsection{Main Capability-Integration Results}
\label{sec:main-results}

\begin{figure}[t]
    \centering
    \captionsetup[subfigure]{font=small,labelfont=normalfont,textfont=normalfont}
    \includegraphics[width=\columnwidth]{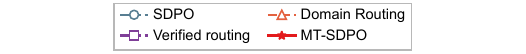}

    \begin{subfigure}[t]{0.49\columnwidth}
        \centering
        \includegraphics[width=\textwidth]{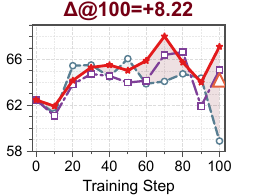}
        \caption{Macro Accuracy $\uparrow$}
    \end{subfigure}
    \hfill
    \begin{subfigure}[t]{0.49\columnwidth}
        \centering
        \includegraphics[width=\textwidth]{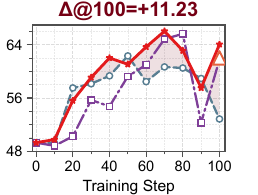}
        \caption{Worst-Domain Accuracy $\uparrow$}
    \end{subfigure}

    \begin{subfigure}[t]{0.49\columnwidth}
        \centering
        \includegraphics[width=\textwidth]{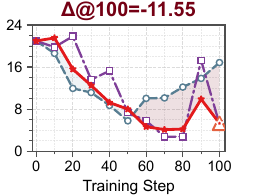}
        \caption{Domain Gap $\downarrow$}
    \end{subfigure}
    \hfill
    \begin{subfigure}[t]{0.49\columnwidth}
        \centering
        \includegraphics[width=\textwidth]{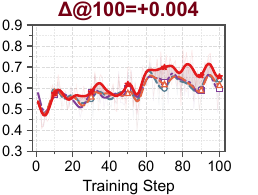}
        \caption{Rollout Accuracy $\uparrow$}
    \end{subfigure}

    \caption{Held-Out Dynamics and Rollout Accuracy. Panels (a)--(c) are held-out metrics; (d) is the verifier pass rate on training rollouts. Shading is the signed \mbox{\OURS--SDPO} difference, not a confidence interval, and Domain Routing appears only at step 100 in (a)--(c).}
    \label{fig:heldout-training-dynamics}
\end{figure}

\paragraph{\OURS recovers the weakest domain of Qwen3-8B.} Table~\ref{tab:main-results} and Figure~\ref{fig:main-endpoint} report every method at step 100. On Qwen3-8B, \OURS improves Macro and Worst-domain accuracy by 4.64 and 14.79 points over Multi-Domain Post-Training, and narrows the gap from 20.96 to 5.30. Chemistry, the weakest domain, rises 18.59 points, while materials and physics together give back under five. \OURS integrates capability instead of improving every domain, which is what one deployed policy needs.

\paragraph{On Qwen3-8B, one integrated policy is more balanced than one matched teacher per domain.} \OURS improves on the three-model matched-teacher reference by 3.79 Macro and 11.15 Worst-domain points with a single model at inference. Domain Routing reaches the same 5.30 gap but stays 2.90 and 2.05 points lower on the two accuracy metrics: routing alone recovers the balance, and verification and aggregation supply the accuracy.

\paragraph{Spending the verifier on allocation is what redistributes capability.} Spent as a scalar reward instead, the signal varies the objective, not the allocation rule. On Qwen3-8B that GRPO reference widens the gap from 20.96 to 21.76, whereas \OURS brings the same start to 5.30 and stands 9.17 points higher in its weakest domain. Appendix~\ref{sec:appendix-grpo} gives its setup and per-domain scores.

\paragraph{A flatter profile arises through two different mechanisms.} All three Qwen3 scales cut the domain gap, by 49.6\% at 1.7B, 41.5\% at 4B, and 74.7\% at 8B, but Macro improves at 8B alone. At 1.7B and 4B the flattening comes mostly from the strong domains falling: the weakest domain gains 4.19 and 1.97 points while the other two give back 11.86 and 13.85. At 8B and on OLMo-3-7B-Instruct, the weakest domain instead rises by 18.59 and 7.27 points and carries Macro with it, and OLMo gives the best Macro, Worst, and Gap of its block. Verified feedback therefore levels the profile wherever headroom remains, but yields an aggregate gain only where the weakest domain itself improves.

\begin{table}[t]
\centering
\small
\setlength{\tabcolsep}{1pt}
\newcommand{\abldelta}[1]{\mbox{$(#1)$}}
\begin{tabular*}{\columnwidth}{@{\extracolsep{\fill}}lrrr@{}}
\toprule
Method & Macro $\uparrow$ ($\Delta$) & Worst $\uparrow$ ($\Delta$) & Gap $\downarrow$ ($\Delta$) \\
\midrule
w/o Cross-Domain & 65.08\abldelta{+2.61} & 61.79\abldelta{+12.50} & \textbf{5.14}\abldelta{-15.82} \\
w/o Aggregation & \underline{66.11}\abldelta{+3.64} & \underline{62.26}\abldelta{+12.97} & 7.99\abldelta{-12.97} \\
w/o Verification & 60.55\abldelta{-1.92} & 55.14\abldelta{+5.85} & 10.92\abldelta{-10.04} \\
\textbf{\OURS{}} & \textbf{67.11}\abldelta{+4.64} & \textbf{64.08}\abldelta{+14.79} & \underline{5.30}\abldelta{-15.66} \\
\bottomrule
\end{tabular*}
\caption{Ablation of Supervision Allocation on Qwen3-8B. $\Delta$ is method minus the shared Multi-Domain PT initialization; bold and underline mark the best and second-best step-100 values per column.}
\label{tab:ablations}
\end{table}

\paragraph{Headroom in the weakest domain decides whether the online phase pays off.} Multi-domain post-training alone lifts Llama-3.1-8B-Instruct by 21.04 Worst-domain points and leaves it the flattest initialization of the five, so the weakest-domain room that \OURS exploits elsewhere is spent before the online phase begins and no controlled method improves on it. The bound is the remaining headroom, not an ordering among methods.

\subsection{Training Dynamics}
\label{sec:training-dynamics}

\paragraph{Non-monotonic trajectories motivate the fixed step-100 comparison.} Figure~\ref{fig:heldout-training-dynamics} places the held-out integration metrics and the training-time rollout accuracy in one view. SDPO peaks at step 50 and loses more than seven Macro points by step 100, whereas \OURS peaks at step 70 and holds most of that peak to the end. Because no method improves monotonically, we compare at step 100 rather than at each method's best intermediate checkpoint.

\subsection{Ablation Studies}
\label{sec:ablations}

\paragraph{Cross-domain reassignment and aggregation play complementary roles.} Table~\ref{tab:ablations} expresses each variant as a removal from full \OURS, with the self-anchor, sanitizer, EMA self-teacher, initialization, and update budget fixed. Restricting feedback to the verified domain-matched teacher (w/o Cross-Domain) starts the ladder; admitting one sample-eligible teacher instead (w/o Aggregation) adds 1.03 Macro and 0.47 Worst-domain points but widens the gap from 5.14 to 7.99; aggregating every eligible teacher in full \OURS adds 1.00 and 1.82 points more and brings the gap to 5.30, recovering the balance.

\paragraph{Verification makes the larger teacher pool useful.} Replacing the eligible set with all three unverified teachers (w/o Verification) loses 6.56 Macro and 8.94 Worst-domain points against full \OURS, and falls 1.92 Macro points below the shared initialization. The offline manifest shows why: the unverified rule draws on an incorrect answer in 50.21\% of teacher--sample pairs, against 47.06\% under domain routing.

\section{Conclusion}
\label{sec:conclusion}

We showed that a teacher pool is better allocated per sample than per domain. Sample-level verification recovers a quarter of the samples domain routing misses; cross-domain reassignment then buys accuracy at the cost of balance, and aggregating every eligible teacher wins both back. Two limits remain: a third of samples admit no correct teacher, and balanced initializations leave little headroom. Extending \OURS beyond exact-answer settings will require a reliable, task-appropriate alternative to exact-match eligibility.

\bibliography{references}

\clearpage
\appendix
\twocolumn[{
\begin{center}
    {\Large\bfseries Appendix}\\[10pt]
\end{center}
}]
\noindent This appendix provides the training, evaluation, and prompt
details behind the main text, together with additional results. It is
organized as follows.
\begin{center}\vspace{-4pt}\rule{0.6\linewidth}{0.4pt}\end{center}
{\small
\noindent\textbf{\ref{sec:appendix-post-training}\quad Post-Training and Initialization}\dotfill\textbf{\pageref{sec:appendix-post-training}}\\[2pt]
\noindent\textbf{\ref{sec:appendix-online-details}\quad Online Optimization and Teacher Feedback}\dotfill\textbf{\pageref{sec:appendix-online-details}}\\[2pt]
\noindent\hspace{1.2em}\ref{sec:appendix-loss-form}\quad Implemented Form of the Distillation Objective\dotfill\pageref{sec:appendix-loss-form}\\[1pt]
\noindent\hspace{1.2em}\ref{sec:appendix-manifest}\quad Private-Answer Manifest\dotfill\pageref{sec:appendix-manifest}\\[1pt]
\noindent\hspace{1.2em}\ref{sec:appendix-feedback-details}\quad Rollout-Specific Feedback\dotfill\pageref{sec:appendix-feedback-details}\\[1pt]
\noindent\textbf{\ref{sec:appendix-evaluation}\quad Evaluation Protocol}\dotfill\textbf{\pageref{sec:appendix-evaluation}}\\[2pt]
\noindent\hspace{1.2em}\ref{sec:appendix-parsing-fallback}\quad Guarded Parsing Fallback\dotfill\pageref{sec:appendix-parsing-fallback}\\[1pt]
\noindent\textbf{\ref{sec:appendix-additional-results}\quad Additional Results}\dotfill\textbf{\pageref{sec:appendix-additional-results}}\\[2pt]
\noindent\hspace{1.2em}\ref{sec:appendix-cross-scale-dynamics}\quad Cross-Scale Training Dynamics\dotfill\pageref{sec:appendix-cross-scale-dynamics}\\[1pt]
\noindent\hspace{1.2em}\ref{sec:appendix-training-diagnostics}\quad Additional Feedback Diagnostics\dotfill\pageref{sec:appendix-training-diagnostics}\\[1pt]
\noindent\hspace{1.2em}\ref{sec:appendix-grpo}\quad Verifiable-Reward RL Reference\dotfill\pageref{sec:appendix-grpo}\\[1pt]
\noindent\textbf{\ref{sec:appendix-prompts}\quad Prompt Templates}\dotfill\textbf{\pageref{sec:appendix-prompts}}\\[2pt]
\noindent\hspace{1.2em}\ref{sec:appendix-sft-construction-prompts}\quad SFT Data Construction\dotfill\pageref{sec:appendix-sft-construction-prompts}\\[1pt]
\noindent\hspace{1.2em}\ref{sec:appendix-student-rollout-prompts}\quad Student SFT and Initial Rollout\dotfill\pageref{sec:appendix-student-rollout-prompts}\\[1pt]
\noindent\hspace{1.2em}\ref{sec:appendix-teacher-reprompts}\quad Online Self-Teacher Reprompts\dotfill\pageref{sec:appendix-teacher-reprompts}\\[1pt]
\noindent\hspace{1.2em}\ref{sec:appendix-feedback-prompt}\quad Frozen Teacher Diagnostic Prompt\dotfill\pageref{sec:appendix-feedback-prompt}\\[1pt]
\noindent\hspace{1.2em}\ref{sec:appendix-evaluation-prompts}\quad Evaluation Prompts\dotfill\pageref{sec:appendix-evaluation-prompts}\\[1pt]
\noindent\textbf{\ref{sec:appendix-feedback-effectiveness}\quad Feedback Effectiveness}\dotfill\textbf{\pageref{sec:appendix-feedback-effectiveness}}\\[2pt]
}
\begin{center}\vspace{-8pt}\rule{0.6\linewidth}{0.4pt}\end{center}
\vspace{6pt}

\section{Post-Training and Initialization}
\label{sec:appendix-post-training}

For five student backbones (Qwen3-1.7B, Qwen3-4B, Qwen3-8B, Llama-3.1-8B-Instruct, and OLMo-3-7B-Instruct), we construct three domain teachers and one Multi-Domain Post-Training policy from the family-matched base model. The teachers use separate domain sets; the Multi-Domain policy uses their balanced union. Post-training, controlled online optimization, and evaluation follow the shared recipes in Tables~\ref{tab:appendix-post-training}--\ref{tab:appendix-evaluation}. Identifiers are family-specific, and family-matched generators and validation/recovery pipelines produce the supervision. The teachers use their final three-epoch weights and remain frozen. Every controlled online method starts from the corresponding epoch-2 Multi-Domain checkpoint; no method-specific initialization is used.

\begin{table}[t]
\centering
\small
\renewcommand{\arraystretch}{1.1}
\setlength{\tabcolsep}{8pt}
\begin{tabularx}{\linewidth}{X l}
\toprule
\textbf{Parameter} & \textbf{Value} \\
\midrule
\multicolumn{2}{l}{\textbf{General}} \\
\midrule
Model update & Full-parameter post-training \\
Optimizer & Fused AdamW \\
Learning rate & $1\times10^{-5}$ \\
Schedule & Cosine, 5\% warmup \\
Training epochs & 3 \\
\addlinespace
\multicolumn{2}{l}{\textbf{Data}} \\
\midrule
Global batch size & 4 \\
Maximum sequence length & 8,192 tokens \\
Response handling & Response tokens only; no packing \\
\addlinespace
\multicolumn{2}{l}{\textbf{Training}} \\
\midrule
Weight decay / gradient clipping & 0.1 / 1.0 \\
Numeric precision & bfloat16 \\
\bottomrule
\end{tabularx}
\caption{Post-Training Settings Shared Across All Student Backbones. Domain-specific and Multi-Domain Post-Training use the same recipe for Qwen3, Llama-3.1, and OLMo-3.}
\label{tab:appendix-post-training}
\end{table}

\section{Online Optimization and Teacher Feedback}
\label{sec:appendix-online-details}

Algorithm~\ref{alg:mt-sdpo} gives the procedure described in Section~\ref{sec:method} of the main text. Table~\ref{tab:appendix-online} lists the online optimization budget and distillation settings. Each controlled method processes 32 prompts per update, draws eight on-policy responses per prompt, and runs for 100 updates. The methods use the same prompt IDs, rollout-sampling configuration, and student-update budget. Each runs on eight H100 80GB GPUs. SDPO allocates all eight GPUs to the policy; methods with external teachers split four for policy training and four for serving the frozen teachers.

\begin{algorithm}[t]
\caption{Answer-Verified MT-SDPO}
\label{alg:mt-sdpo}
\small
\textbf{Input}: data $\mathcal D$; student $\pi_\theta$; EMA self-teacher $\pi_{\bar\theta}$; frozen teachers $\{E_k\}_{k=1}^{K}$; rollout count $G$; EMA rate $\tau$\\
\textbf{Output}: updated student model $\pi_\theta$
\begin{algorithmic}[1]
\STATE \textbf{// Offline: cache sample-level teacher eligibility}
\STATE $\mathcal V_i\leftarrow\{k:\textsc{Verify}(E_k^{\mathrm{ans}}(x_i),a_i^*)=1\},\ \forall(x_i,a_i^*)\in\mathcal D$
\STATE \textbf{// Online: align supervision to each rollout}
\FOR{each online training step}
    \STATE $\mathcal B\sim\mathcal D$; $y_i^{1:G}\sim\pi_\theta(\cdot\mid x_i)$ for each $x_i\in\mathcal B$
    \STATE $\mathcal S_i\leftarrow\{j:\textsc{Verify}(y_i^j,a_i^*)=1\}$
    \FOR{each rollout $y_i^j$ in the batch}
        \STATE $s_i^j\leftarrow\operatorname{Trunc}(\text{earliest rollout in }\mathcal S_i\setminus\{j\})$, or $\varnothing$ \hfill $\boldsymbol{\triangleright}$ Eq.~\eqref{eq:self-anchor}, main text
        \STATE $F_i^j\leftarrow\varnothing$
        \IF{$j\notin\mathcal S_i$}
            \FOR{each $k\in\mathcal V_i$}
                \STATE $F_i^j\leftarrow\operatorname{Concat}\!\left[F_i^j,\Gamma(E_k^{\mathrm{fb}}(x_i,y_i^j))\right]$
            \ENDFOR
        \ENDIF
        \STATE $c_i^j\leftarrow[s_i^j;F_i^j]$; $m_i^j\leftarrow\mathbb 1[c_i^j\neq\varnothing]$
        \STATE Accumulate the $(i,j)$ terms of $\mathcal L_{\mathrm{MT\text{-}SDPO}}$ \hfill $\boldsymbol{\triangleright}$ Eq.~\eqref{eq:mt-sdpo-loss}, main text
    \ENDFOR
    \STATE Update $\theta$ over the collected batch
    \STATE $\bar\theta\leftarrow(1-\tau)\bar\theta+\tau\theta$
\ENDFOR
\STATE \textbf{return} $\pi_\theta$; discard $\pi_{\bar\theta}$ and all teachers
\end{algorithmic}
\end{algorithm}

\subsection{Implemented Form of the Distillation Objective}
\label{sec:appendix-loss-form}

Equation~\eqref{eq:mt-sdpo-loss} of the main text states the objective in its masked, importance-weighted form. Three details are fixed by the implementation. First, the Jensen--Shannon divergence $D_{\mathrm{JSD}}(p\Vert q)=\tfrac12 D_{\mathrm{KL}}(p\Vert u)+\tfrac12 D_{\mathrm{KL}}(q\Vert u)$, where $u=\tfrac12(p+q)$, is evaluated on a truncated support rather than the full vocabulary. At each position, the operator $\mathcal T_{i,t}^{j}[\cdot]$ retains the 100 tokens with the highest student probability and collects the remaining mass in a tail symbol. The student defines the support, while each branch retains its own tail mass. Second, the weight $w_{i,t}^{j}=\min(w_{i,t}^{j,\mathrm{prox}},\rho)\cdot\min(w_{i,t}^{j,\mathrm{TIS}},\rho)$ combines the PPO proximal ratio $w_{i,t}^{j,\mathrm{prox}}=\pi_\theta(y_{i,t}^{j})/\pi_{\theta_{\mathrm{old}}}(y_{i,t}^{j})$ with the truncated-importance-sampling ratio $w_{i,t}^{j,\mathrm{TIS}}=\pi_{\theta_{\mathrm{old}}}(y_{i,t}^{j})/\pi_{\theta_{\mathrm{rollout}}}(y_{i,t}^{j})$. Both ratios are conditioned on $(x_i,y_{i,<t}^{j})$, detached, and clipped only from above at $\rho=2$. Third, the normalizer is applied per micro-batch rather than globally. For dynamic micro-batches $\{\mathcal M_b\}_{b=1}^{B_\mu}$, let $n_b=|\mathcal M_b|$, $N=\sum_b n_b$, and $Z_b=\max\{1,\sum_{(i,j)\in\mathcal M_b,t}m_i^jM_{i,t}^{j}\}$. The implemented loss is the sequence-count-weighted average of the micro-batch token means:
\begin{equation}
\begin{aligned}
\mathcal L_{\mathrm{MT\text{-}SDPO}}
&=\sum_{b=1}^{B_\mu}\frac{n_b}{N}\frac{1}{Z_b}
\sum_{\substack{(i,j)\in\mathcal M_b\\ t}}
m_i^jM_{i,t}^{j}w_{i,t}^{j}\\[-0.2em]
&\quad\cdot D_{\mathrm{JSD}}\!\left(
\mathcal T_{i,t}^{j}\!\left[p_{\theta,i,t}^{j}\right]\middle\Vert
\mathcal T_{i,t}^{j}\!\left[q_{\bar\theta,i,t}^{j}\right]\right).
\end{aligned}
\label{eq:appendix-mt-sdpo-loss}
\end{equation}

\begin{table}[t]
\centering
\small
\renewcommand{\arraystretch}{1.1}
\setlength{\tabcolsep}{8pt}
\begin{tabularx}{\linewidth}{X l}
\toprule
\textbf{Parameter} & \textbf{Value} \\
\midrule
\multicolumn{2}{l}{\textbf{Training}} \\
\midrule
Prompts per update / rollouts per prompt & 32 / 8 \\
Number of updates & 100 \\
Student learning rate & $1\times10^{-5}$ \\
\addlinespace
\multicolumn{2}{l}{\textbf{Batching / Rollout}} \\
\midrule
Rollout sampling & Temperature $1.0$, top-$p=1.0$ \\
Maximum prompt length & 2,048 tokens \\
Maximum response length & 8,192 tokens \\
Optimization mini-batch & 32, dynamic micro-batching \\
Dynamic token cap & 18,944 tokens per GPU \\
\addlinespace
\multicolumn{2}{l}{\textbf{Distillation}} \\
\midrule
Distillation support & Student top-100 tokens plus tail mass \\
JSD mixture coefficient & Fixed at $1/2$ \\
EMA update rate & $\tau=0.05$ per rollout batch \\
Importance-ratio upper clip & 2.0 for both ratios \\
\addlinespace
\multicolumn{2}{l}{\textbf{Feedback processing and gating}} \\
\midrule
Anchor token budget & 4,096 \\
Feedback token budget & 1,024 \\
Maximum privileged re-prompt length & 10,240 tokens \\
\bottomrule
\end{tabularx}
\caption{Online Settings Shared Across Student Families and Controlled Methods. The listed settings are shared; the methods differ in their available privileged context as defined in the main text.}
\label{tab:appendix-online}
\end{table}

\subsection{Private-Answer Manifest}
\label{sec:appendix-manifest}

For every student family, each frozen teacher answers each of the 1{,}953 training questions once before online training. Decoding is deterministic, with temperature $0$, top-$p=1.0$, and a maximum of 6,144 tokens. Manifest construction uses the generic SciKnowEval L3 rollout prompt reproduced in Section~\ref{sec:appendix-student-rollout-prompts}. This prompt is also used for online training and evaluation. The programmatic answer checker compares the parsed answer with the reference answer and records the sample-level eligible set. Neither the private-answer prompt nor the subsequent teacher-feedback prompt contains the reference answer. MT-SDPO queries every eligible teacher. The w/o Cross-Domain control (verified domain routing) restricts this set to the nominal-domain teacher. Domain Routing and w/o Verification (the unverified all-teacher control) do not use the manifest for eligibility.

\subsection{Rollout-Specific Feedback}
\label{sec:appendix-feedback-details}

External feedback is requested only when the student rollout is incorrect. Eligible teachers are queried independently in the fixed order of chemistry, materials science, and physics. Decoding uses temperature $0.2$, top-$p=0.95$, and a 256-token output limit. A message is retained only if it has a valid concept anchor under the lexical sanitizer and passes the answer-leakage checks. Failed or rejected messages are dropped. The retained messages are concatenated with the available self-anchor under the token budgets in Table~\ref{tab:appendix-online}.

\section{Evaluation Protocol}
\label{sec:appendix-evaluation}

All five student backbones and all reported checkpoints use the same frozen evaluation protocol. Each domain contains 79 held-out questions, and each question receives 16 independently sampled responses. Table~\ref{tab:appendix-evaluation} records the settings that affect the reported avg@16 results.

\begin{table}[t]
\centering
\small
\renewcommand{\arraystretch}{1.1}
\setlength{\tabcolsep}{8pt}
\begin{tabularx}{\linewidth}{X l}
\toprule
\textbf{Parameter} & \textbf{Value} \\
\midrule
\multicolumn{2}{l}{\textbf{Sampling}} \\
\midrule
Responses per question & 16 \\
Sampling & Temperature $0.6$, top-$p=0.95$, top-$k=20$ \\
Maximum new tokens & 6,144 \\
\addlinespace
\multicolumn{2}{l}{\textbf{Evaluation}} \\
\midrule
Primary parser & Last explicit \texttt{answer} field \\
Missing, invalid, or truncated answer & Incorrect \\
Macro aggregation & Unweighted mean over three domains \\
\bottomrule
\end{tabularx}
\caption{Frozen Evaluation Settings Shared Across All Student Backbones}
\label{tab:appendix-evaluation}
\end{table}

\subsection{Guarded Parsing Fallback}
\label{sec:appendix-parsing-fallback}

When the primary parser does not establish a correct answer, a fixed Qwen3-8B base model is queried as a deterministic binary judge (temperature $0$, top-$p=1.0$). The fallback cannot overturn an explicitly extracted non-gold answer, and a judge parse failure is counted as incorrect. This rule handles ambiguous formatting without allowing the judge to replace the answer key.

\section{Additional Results}
\label{sec:appendix-additional-results}

\subsection{Cross-Scale Training Dynamics}
\label{sec:appendix-cross-scale-dynamics}

\begin{figure*}[t]
    \centering
    \includegraphics[width=\textwidth]{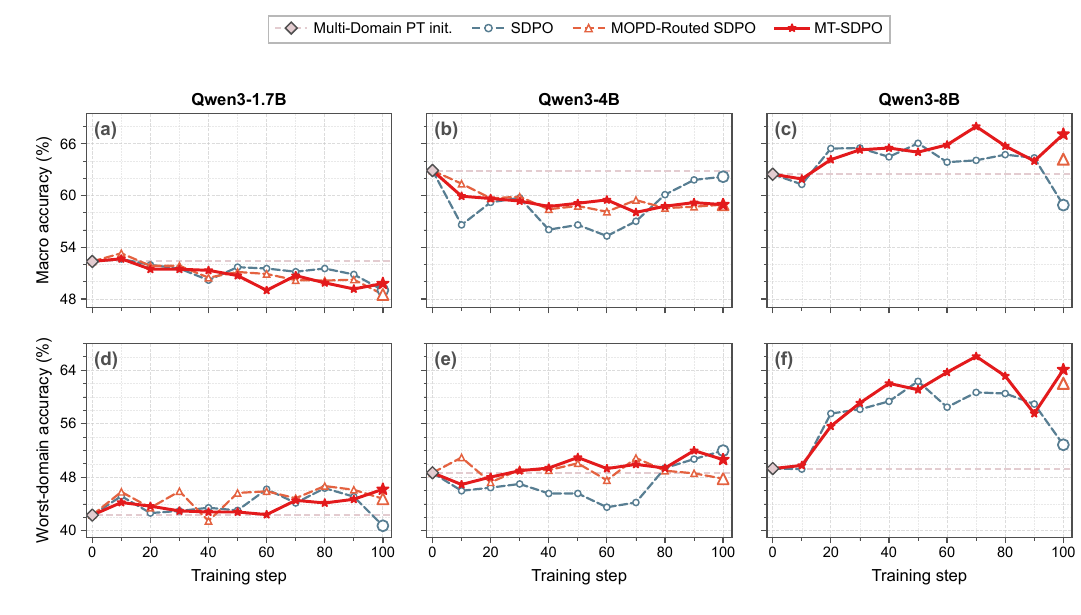}
    \caption{Cross-Scale Held-Out Dynamics, Evaluated Every 10 Training Steps. Columns are Qwen3-1.7B, Qwen3-4B, and Qwen3-8B, and the two rows report Macro and Worst-domain accuracy. The horizontal reference marks each student's Multi-Domain Post-Training initialization, and lines connect evaluated checkpoints without smoothing. Domain Routing has a held-out result only at step 100 on Qwen3-8B, so it appears as a single step-100 marker in (c) and (f). The legend label MOPD-Routed SDPO denotes Domain Routing in the main text.}
    \label{fig:appendix-cross-scale-dynamics}
\end{figure*}

Figure~\ref{fig:appendix-cross-scale-dynamics} reports the held-out Macro and Worst-domain trajectories for every completed Qwen3 student scale. Intermediate checkpoints are diagnostic only: the main comparison continues to use the common, pre-specified step-100 checkpoint for every online method.

The trajectories differ across scales. At step 100, Qwen3-1.7B and Qwen3-4B primarily recover weakest-domain capability rather than improving Macro accuracy over their respective initializations. On Qwen3-8B, MT-SDPO finishes above its initialization and SDPO on both Macro and Worst-domain accuracy. The non-monotonic trajectories support the use of a common checkpoint rather than method-specific checkpoint selection.

\subsection{Additional Feedback Diagnostics}
\label{sec:appendix-training-diagnostics}

\begin{figure*}[t]
    \centering
    \captionsetup[subfigure]{font=normalsize,labelfont=normalfont,textfont=normalfont}
    \includegraphics[width=0.84\textwidth]{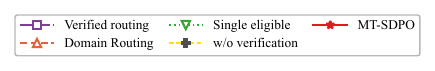}

    \begin{subfigure}[t]{0.322\textwidth}
        \centering
        \includegraphics[width=\textwidth]{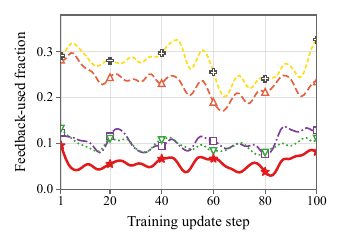}
        \caption{Feedback Use Trend}
    \end{subfigure}
    \hfill
    \begin{subfigure}[t]{0.322\textwidth}
        \centering
        \includegraphics[width=\textwidth]{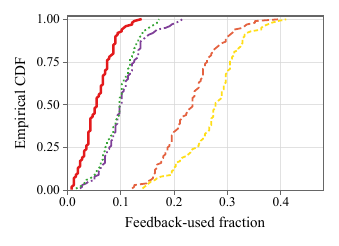}
        \caption{Feedback Use Distribution}
    \end{subfigure}
    \hfill
    \begin{subfigure}[t]{0.322\textwidth}
        \centering
        \includegraphics[width=\textwidth]{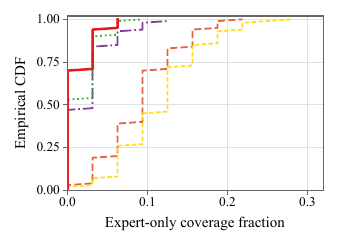}
        \caption{Teacher-Only Coverage}
    \end{subfigure}

    \caption{Realized Teacher-Feedback Exposure on Qwen3-8B over 100 Training Updates. (a) Smoothed fraction of rollouts that use retained teacher feedback. (b) Empirical distribution of the corresponding raw per-update fractions. (c) Empirical distribution of prompts with no successful student rollout but retained teacher feedback. Higher exposure alone implies neither more accurate feedback nor better held-out performance. In the legend, Verified routing, Single eligible, and w/o verification correspond to w/o Cross-Domain, w/o Aggregation, and w/o Verification in the main text, respectively.}
    \label{fig:appendix-feedback-diagnostics}
\end{figure*}

Figure~\ref{fig:appendix-feedback-diagnostics} reports realized feedback exposure on Qwen3-8B. These diagnostics are distinct from the offline teacher-pool coverage and the held-out and rollout-accuracy dynamics in the main text. Offline coverage asks whether a correct teacher exists for a training question. Here, feedback use is measured after the student-error condition, each method's eligibility rule, feedback generation, and sanitization. Panel (a) uses centered Gaussian smoothing ($\sigma=2$) only for visualization. Panels (b) and (c) report empirical distributions over all 100 recorded updates.

Across the 100 updates, Domain Routing and w/o Verification use retained teacher feedback on 23.07\% and 27.32\% of rollouts on average. The corresponding averages are 10.39\% for w/o Cross-Domain, 9.63\% for w/o Aggregation, and 5.64\% for \mbox{MT-SDPO}. Average teacher-only prompt coverage follows the same broad ordering. Domain Routing and w/o Verification reach 9.16\% and 11.66\%, while w/o Cross-Domain, w/o Aggregation, and \mbox{MT-SDPO} reach 2.44\%, 1.81\%, and 1.12\%. This exposure ordering differs from the step-100 performance ordering in the main text. Feedback volume alone therefore does not explain integration quality. In these runs, answer verification is associated with less frequent, more selective external supervision.

The teacher-only metric uses all prompts as its denominator. It is the joint frequency of a missing successful student rollout and available retained teacher feedback. It is not the conditional probability of receiving feedback given a missing self-anchor. Neither metric tests whether the retained feedback is semantically correct or repairs the student's answer. Performance claims remain anchored to the step-100 checkpoint and exact held-out trajectories reported in the main text.

\subsection{Verifiable-Reward RL Reference}
\label{sec:appendix-grpo}

The main text's ablation comparison holds the teacher pool, self-anchor, sanitizer, privileged self-teacher, initialization, and update budget fixed while varying the supervision-allocation rule. GRPO is outside that comparison because it changes the objective and uses no teacher pool. It replaces the token-level distillation objective of Eq.~\eqref{eq:mt-sdpo-loss} with a policy gradient on a scalar reward. The programmatic verifier used by MT-SDPO as an eligibility gate supplies that reward. Table~\ref{tab:appendix-grpo} reports GRPO only as a separate objective reference under the same update and rollout budget.

Each GRPO reference run starts from the corresponding epoch-2 Multi-Domain Post-Training checkpoint used by the controlled methods. It runs for 100 updates with 8 on-policy rollouts per prompt and uses eight H100 80GB GPUs. The verifier and evaluation protocol in Section~\ref{sec:appendix-evaluation} remain the same. No teacher, feedback, or self-teacher component is active, and the reward is the binary verifier outcome on the student's own rollouts.

\begin{table}[t]
\centering
\small
\setlength{\tabcolsep}{3pt}
\begin{tabular*}{\columnwidth}{@{\extracolsep{\fill}}lrrrrrr@{}}
\toprule
Method & Chem. & Mat. & Phys. & Macro & Worst & Gap \\
\midrule
\multicolumn{7}{@{}l}{\textit{Qwen3-1.7B}} \\
Multi-Domain PT & 42.33 & 63.05 & 51.66 & 52.35 & 42.33 & 20.73 \\
GRPO & 45.73 & 64.08 & 54.43 & 54.75 & 45.73 & 18.35 \\
MT-SDPO & 46.52 & 56.65 & 46.20 & 49.79 & 46.20 & 10.44 \\
\addlinespace
\multicolumn{7}{@{}l}{\textit{Qwen3-4B}} \\
Multi-Domain PT & 48.66 & 71.36 & 68.67 & 62.90 & 48.66 & 22.71 \\
GRPO & 52.53 & 74.53 & 66.22 & 64.43 & 52.53 & 21.99 \\
MT-SDPO & 50.63 & 62.26 & 63.92 & 58.94 & 50.63 & 13.29 \\
\addlinespace
\multicolumn{7}{@{}l}{\textit{Qwen3-8B}} \\
Multi-Domain PT & 49.29 & 70.25 & 67.88 & 62.47 & 49.29 & 20.96 \\
GRPO & 54.91 & 76.66 & 70.25 & 67.27 & 54.91 & 21.76 \\
MT-SDPO & 67.88 & 69.38 & 64.08 & 67.11 & 64.08 & 5.30 \\
\addlinespace
\multicolumn{7}{@{}l}{\textit{Llama-3.1-8B-Instruct}} \\
Multi-Domain PT & 56.33 & 59.57 & 51.42 & 55.78 & 51.42 & 8.15 \\
GRPO & 70.65 & 66.46 & 64.16 & 67.09 & 64.16 & 6.49 \\
MT-SDPO & 59.57 & 53.48 & 44.62 & 52.56 & 44.62 & 14.95 \\
\addlinespace
\multicolumn{7}{@{}l}{\textit{OLMo-3-7B-Instruct}} \\
Multi-Domain PT & 48.42 & 63.21 & 57.67 & 56.43 & 48.42 & 14.79 \\
GRPO & 47.47 & 55.85 & 36.87 & 46.73 & 36.87 & 18.98 \\
MT-SDPO & 55.69 & 65.82 & 64.55 & 62.02 & 55.69 & 10.13 \\
\bottomrule
\end{tabular*}
\caption{Verifiable-Reward RL Reference from the Shared Initialization. Columns report step-100 avg@16 accuracy (\%) under the evaluation protocol of Section~\ref{sec:appendix-evaluation}. Within each block, Multi-Domain Post-Training is the shared initialization for GRPO and MT-SDPO. GRPO changes the training objective and is reported outside the main text's controlled online-distillation comparison.}
\label{tab:appendix-grpo}
\end{table}

The GRPO results are reported descriptively across all five backbones. Because GRPO changes the training objective, these results are separate from the main text's controlled supervision-allocation comparison and do not establish a general ordering between the two objectives.

\section{Prompt Templates}
\label{sec:appendix-prompts}

This section reports the exact message text used for SciKnowEval L3 data construction, post-training, online training, and evaluation. The SFT construction prompts in Section~\ref{sec:appendix-sft-construction-prompts} document the Qwen3 pipeline. Llama and OLMo use family-matched generators and their corresponding validation/retry pipelines rather than reusing Qwen3 responses. The student rollout, online self-teacher reprompt, teacher-feedback, and evaluation instructions are shared across families at the natural-language level. Tokenization and chat-template handling remain model-specific. Braced fields denote runtime substitutions rather than literal input. The field \texttt{\{choices\_block\}} renders each option in its original order as \texttt{\{label\}. \{text\}}. The domain-specific field \texttt{\{domain\_expert\}} is \texttt{chemistry expert}, \texttt{materials science expert}, or \texttt{physics expert}. The field \texttt{\{expert\_domain\}} is the corresponding display name.

\subsection{SFT Data Construction}
\label{sec:appendix-sft-construction-prompts}

\subsubsection{Independent Generator Sampling}

Two independent candidates are sampled for each training record. The reference answer is not included in this prompt.

\begin{promptbox}{Independent generator sampling}
\begin{lstlisting}[style=promptstyle]
System:
You are a knowledgeable and rigorous {domain_expert}.

User:
Solve the following multiple-choice question. Give a concise, self-contained derivation using the relevant scientific principles, equations, calculations, or evidence. Include enough explicit reasoning for a reviewer to check, but avoid unnecessary background, repeated calculations, speculative alternatives, or restating the problem. Do not return only an answer.

Question:
{question}

Choices:
{choices_block}

End your response with exactly one final line in this format: "answer": "C". Replace C with your selected choice letter. Do not add Markdown styling, a code fence, a heading, or any text after that final line.
\end{lstlisting}
\end{promptbox}

\subsubsection{Gold-Conditioned Fallback}

This fallback is used only if neither independent candidate yields a parseable correct label. Failed independent completions are not inserted into the fallback prompt.

\begin{promptbox}{Gold-conditioned fallback generation}
\begin{lstlisting}[style=promptstyle]
System:
You are a knowledgeable and rigorous {domain_expert}. You may receive a private correctness constraint. Use it only internally, and never mention or allude to its existence in your response.

User:
Construct a concise, self-contained solution to the following multiple-choice question. Use the private correctness constraint internally and build a scientifically valid derivation that genuinely supports it. Include enough explicit reasoning for a reviewer to check, but avoid unnecessary background, repeated calculations, speculative alternatives, or restating the problem. Do not return only an answer.

Never mention or allude to the private constraint, an answer key, a hint, or any supplied answer. If the constraint cannot be scientifically justified, explain only the scientific inconsistency in the problem without disclosing that a constraint was supplied.

Question:
{question}

Choices:
{choices_block}

Private correctness constraint:
{answer_key}. {answer_text}

End your response with exactly one final line in this format: "answer": "C". Replace C with the constrained choice letter. Do not add Markdown styling, a code fence, a heading, or any text after that final line.
\end{lstlisting}
\end{promptbox}

\subsubsection{Self-Verifier}

\begin{promptbox}{Self-verifier for generated reasoning}
\begin{lstlisting}[style=promptstyle]
System:
You are a rigorous reviewer of scientific reasoning.

User:
Review the proposed solution to the following multiple-choice question.

Check whether its scientific facts, equations, calculations, units, and logical steps are correct. A correct final choice is not sufficient: reject the solution if it reaches the correct answer through incorrect, unsupported, circular, or fabricated reasoning.

Do not rewrite the solution. Briefly explain your judgment and finish with exactly one of the following plain-text lines. Do not apply Markdown styling to the verdict line.

Verdict: ACCEPT
Verdict: REJECT
Verdict: REVIEW

Question:
{question}

Choices:
{choices_block}

Reference answer:
{answer_key}. {answer_text}

Proposed solution:
{cot_response}
\end{lstlisting}
\end{promptbox}

For the Qwen3 data, the generator and verifier are both Qwen3-32B, so this stage is self-verification rather than independent verification. The retained completion becomes the SFT assistant target without summarization, rewriting, or answer normalization.

\subsection{Student SFT and Initial Rollout}
\label{sec:appendix-student-rollout-prompts}

Student SFT uses a domain-specific system message. Online rollout, offline private-answer manifest construction, and evaluation instead use the generic system message below. Their user message is identical.

\begin{promptbox}{System messages for SFT and rollout}
\begin{lstlisting}[style=promptstyle]
Student SFT system:
You are a knowledgeable and rigorous {domain_expert}.

Online rollout, manifest, and evaluation system:
You are a knowledgeable and rigorous scientific reasoning assistant.
\end{lstlisting}
\end{promptbox}

\begin{promptbox}{Shared multiple-choice user message}
\begin{lstlisting}[style=promptstyle]
Solve the following multiple-choice question. Give a concise, self-contained derivation using the relevant scientific principles, equations, calculations, or evidence. Include enough explicit reasoning for a reviewer to check, but avoid unnecessary background, repeated calculations, speculative alternatives, or restating the problem. Do not return only an answer.

Question:
{question}

Choices:
{choices_block}

Please show your choice in the answer field with only the choice letter, e.g., "answer": "C".
\end{lstlisting}
\end{promptbox}

\subsection{Online Self-Teacher Reprompts}
\label{sec:appendix-teacher-reprompts}

The self-teacher reprompt is assembled from the original rollout prompt and independently available context blocks. An unavailable solution or feedback source contributes an empty string.

\subsubsection{SDPO}

\begin{promptbox}{SDPO self-teacher reprompt}
\begin{lstlisting}[style=promptstyle]
Main template:
{prompt}{solution}{feedback}

Correctly solve the original question.

Successful solution section:

Correct solution:

{successful_previous_attempt}

Environment feedback section:

The following is feedback from your unsuccessful earlier attempt:

{feedback_raw}
\end{lstlisting}
\end{promptbox}

The formal SDPO baseline sets environment feedback to disabled, so the feedback section above is not inserted. The successful solution section is inserted only when a reward-verified successful attempt is available.

\subsubsection{MT-SDPO and Routed Variants}

\begin{promptbox}{MT-SDPO self-teacher reprompt}
\begin{lstlisting}[style=promptstyle]
Main template:
{prompt}{solution}{feedback}

Use the available information to reconsider the reasoning and correctly
solve the original question. Follow the original response format and end
with an answer field containing only the selected choice letter.

Successful solution section:

A reward-verified successful attempt previously generated by the current
target policy is shown below:

{successful_previous_attempt}

Frozen-expert feedback section:

The following diagnostic feedback was produced by frozen domain
specialists for the current attempt:

{feedback_raw}
\end{lstlisting}
\end{promptbox}

MT-SDPO, Domain Routing, and the w/o Cross-Domain, w/o Aggregation, and w/o Verification variants use the same reprompt text. The three ablations correspond to verified domain routing, one eligible teacher, and unverified all-teacher feedback, respectively. They differ only in teacher selection and correctness filtering. The anchor, feedback, and total reprompt budgets are 4,096, 1,024, and 10,240 tokens, respectively.

\subsection{Frozen Teacher Diagnostic Prompt}
\label{sec:appendix-feedback-prompt}

Each selected teacher receives the question, all choices, and the current student response. The concept tag is used for anchor filtering and is removed before retained paragraphs enter the self-teacher's privileged context. Retained outputs are concatenated in the fixed order of chemistry, materials science, and physics.

\begin{promptbox}{Frozen teacher diagnostic}
\begin{lstlisting}[style=promptstyle]
System:
You are a knowledgeable and rigorous scientific reasoning assistant.

User:
You are acting as a frozen {expert_domain} specialist. Diagnose one
response produced by the target model without revealing the final answer.

Complete two steps.

Step 1 — Problem anchor:
Identify one technical term or short concept, containing 1–3 words, that
is central to the question. Copy the wording directly from the question.
Output it first inside <concept>...</concept>. Do not use generic words
such as problem, question, answer, value, calculation, formula, model,
student, expert, or response.

Step 2 — Diagnostic feedback:
After the concept tag, write a short paragraph of 3–5 sentences and no
more than approximately 80 words. Focus on the aspects relevant to your
{expert_domain} expertise when useful. Identify the specific step,
assumption, or concept in the target response that is correct, incorrect,
or missing, and explain how the reasoning should be reconsidered.

Do not solve the entire problem from scratch.
Do not reveal the correct choice letter.
Do not quote the correct option text.
Do not write a final answer.
Do not emit an "answer": "X" field.
Do not mention that an answer key or verifier exists.

Question:
{question}

Choices:
{choices_block}

Target response:
{response}

Output:
<concept>...</concept>
\end{lstlisting}
\end{promptbox}

\noindent\textbf{Decoding:} temperature~$=$~0.2, top-$p$~$=$~0.95, and maximum new tokens~$=$~256.

\subsection{Evaluation Prompts}
\label{sec:appendix-evaluation-prompts}

Candidate generation uses exactly the generic system message and shared user message in Section~\ref{sec:appendix-student-rollout-prompts}. Any operational prefix inserted by a family's chat template is not part of the natural-language instruction or the generated completion.

The fallback judge is called only when the primary regex parser has not established correctness. It cannot overturn an explicitly extracted non-gold answer or accept a truncated response with no answer.

\begin{promptbox}{Guarded fallback judge}
\begin{lstlisting}[style=promptstyle]
System:
You are a strict multiple-choice answer verifier. Ignore all reasoning, compare only the candidate's final selected option with the reference answer, and output only true or false.

User:
Candidate response:
<candidate_response>
{candidate_response}
</candidate_response>

Reference answer:
{answer_key}. {answer_text}

The reference answer label is `{answer_key}`. Ignore all reasoning. Output exactly `true` only if the candidate's last explicit `answer` field or `Final Answer` is also `{answer_key}`. Output exactly `false` for any other answer or no final answer. Do not output any explanation.
\end{lstlisting}
\end{promptbox}

\section{Feedback Effectiveness}
\label{sec:appendix-feedback-effectiveness}

We separate feedback exposure from feedback effectiveness. We compare the exposure statistics in Figure~\ref{fig:appendix-feedback-diagnostics} with the held-out step-100 results in the main text. Methods using more retained feedback do not necessarily score higher. This descriptive mismatch means that feedback volume is not itself evidence of utility. The held-out comparisons in the main text measure the end-to-end outcome of each training method. They do not isolate the causal contribution of an individual feedback message.

The fixed held-out protocol supports a method-level comparison, not a message-level causal repair rate. A direct estimate requires post-feedback completions or paired runs with and without feedback on the same erroneous rollouts.

\end{document}